\documentclass{article}

\usepackage{arxiv}

\usepackage{amsmath,amsfonts,bm}

\def\eqref#1{equation~\ref{#1}}

\def\1{\bm{1}}

\DeclareMathAlphabet{\mathsfit}{\encodingdefault}{\sfdefault}{m}{sl}
\SetMathAlphabet{\mathsfit}{bold}{\encodingdefault}{\sfdefault}{bx}{n}

\usepackage{hyperref}
\usepackage{url}
\usepackage{multirow}
\usepackage{graphicx}
\usepackage{caption}
\usepackage{subcaption}
\usepackage{cleveref}
\usepackage{mathtools}
\usepackage[utf8]{inputenc} 
\usepackage[T1]{fontenc}    
\usepackage{hyperref}       
\usepackage{url}            
\usepackage{booktabs}       
\usepackage{amsfonts}       
\usepackage{nicefrac}       
\usepackage{microtype}      
\usepackage{lipsum}
\usepackage{fancyhdr}       
\usepackage{graphicx}       
\usepackage[numbers, sort, comma, square]{natbib}
\usepackage{filecontents}
\usepackage[]{algorithm}
\usepackage[]{algpseudocode}
\usepackage{amsmath,amssymb,mathtools}
\DeclarePairedDelimiter{\norm}{\lVert}{\rVert}
\graphicspath{{media/}}     

\title{Revisiting Structured Dropout}

\author{
  Yiren Zhao \\
  Imperial College London \\ 
	and University of Cambridge \\
  \texttt{a.zhao@imperial.ac.uk} \\
  \And
  Oluwatomisin Dada\\
  University of Cambridge \\
  \texttt{oluwatomisin.dada@cl.cam.ac.uk} \\
	\And
	Xitong Gao \\
	SIAT \\
  \texttt{xt.gao@siat.ac.cn} \\
	\And
	Robert D Mullins \\
  University of Cambridge \\
	\texttt{robert.mullins@cl.cam.ac.uk} 
}

\begin{document}
\maketitle
\begin{abstract}
	Large neural networks are often overparameterised and prone to overfitting, Dropout is a widely used regularization technique to combat overfitting and improve model generalization. However, unstructured Dropout is not always effective for specific network architectures and this has led to the formation of multiple structured Dropout approaches to improve model performance and, sometimes, reduce the computational resources required for inference. In this work, we revisit structured Dropout comparing different Dropout approaches to natural language processing and computer vision tasks for multiple state-of-the-art networks. Additionally, we devise an approach to structured Dropout we call \textbf{\emph{ProbDropBlock}} which drops contiguous blocks from feature maps with a probability given by the normalized feature salience values. We find that with a simple scheduling strategy the proposed approach to structured Dropout consistently improved model performance compared to baselines and other Dropout approaches on a diverse range of tasks and models. In particular, we show \textbf{\emph{ProbDropBlock}} improves RoBERTa finetuning on MNLI by $0.22\%$, and training of ResNet50 on ImageNet by $0.28\%$. 
	
	\end{abstract}
	
	\section{Introduction}
In our modern society, Deep Neural Networks have become increasingly ubiquitous, having achieved significant success in many tasks including visual recognition and natural language processing \cite{heaton2020applications,jumper2021highly,schrittwieser2020mastering}. These networks now play a larger role in our lives and our devices, however, despite their successes they still have notable weaknesses. Deep Neural Networks are often found to be highly overparameterized, and as a result, require excessive memory and significant computational resources. Additionally, due to overparameterization, these networks are prone to overfit their training data. 

There are several approaches to mitigate overfitting including reducing model size or complexity, early stopping \cite{caruana2000overfitting}, data augmentation \citep{devries2017improved} and regularisation \citep{loshchilov2017decoupled}. In this paper, we focus on Dropout which is a widely used form of regularisation proposed by \citet{srivastava2014dropout}. Standard Unstructured Dropout involves randomly deactivating a subset of neurons in the network for each training iteration and training this subnetwork, at inference time the full model could then be treated as an approximation of an ensemble of these subnetworks. 

Unstructured Dropout was efficient and effective and this led to it being widely adopted, however, when applied to Convolutional Neural Networks (CNNs), unstructured Dropout struggled to achieve notable improvements \cite{he2016deep,huang2017densely} and this led to the development of several structured Dropout approaches \cite{ghiasi2018dropblock,dai2019batch,cai2019effective} including DropBlock and DropChannel. DropBlock considers the spatial correlations between nearby entries in a feature map of a CNN and attempts to stop that information flow by deactivating larger contiguous areas/blocks, while DropChannel considers the correlation of information within a particular channel and performs Dropout at the channel level. However, since the development of these structured approaches, there have been further strides in network architecture design, with rising spread and interest in Transformer-based models. 
 
Given the success achieved by block-wise structured Dropout on CNNs, it is only natural to ask the question, \emph{do these approaches apply to Transformer-based models?} 
Structured Dropout approaches for transformers seem to focus on reducing the model size and inference time, these works place more emphasis on pruning or reducing computational resources \cite{xin2020deebert,fan2019reducing} than combating overfitting which is the focus of this paper. 

In this paper, we revisit the idea of structured Dropout for current state-of-the-art models on language and vision tasks. Additionally, we devised our own form of adaptive structured Dropout - \textbf{\emph{ProbDropBlock}} and compare it to preexisting approaches to structured and unstructured Dropout.

\begin{figure}[!t]
	\centering
	\begin{subfigure}{.8\textwidth}
		\centering
		\includegraphics[width=0.95\linewidth]{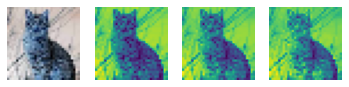}
		\caption{Original image and its RGB channels}
		\label{fig:orignal}
	\end{subfigure}%
	\medskip

	\begin{subfigure}{.46\textwidth}
		\centering
		\includegraphics[width=0.9\linewidth]{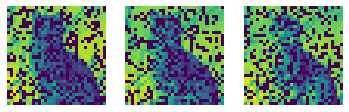}
		\caption{Dropout}
		\label{fig:dropout}
	\end{subfigure}%
	\begin{subfigure}{.46\textwidth}
		\centering
		\includegraphics[width=0.9\linewidth]{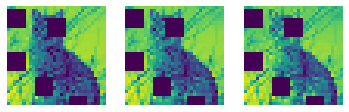}
		\caption{BatchDropBlock}
		\label{fig:dropblock}
	\end{subfigure}
	\begin{subfigure}{.46\textwidth}
		\centering
		\includegraphics[width=0.9\linewidth]{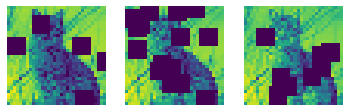}
		\caption{DropBlock}
		\label{fig:channel}
	\end{subfigure}%
	\begin{subfigure}{.46\textwidth}
		\centering
		\includegraphics[width=0.9\linewidth]{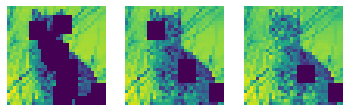}
		\caption{Adaptive DropBlock}
		\label{fig:adaptive}
	\end{subfigure}

	\caption{An illustration of applying different Dropouts to an image.}
	\label{fig:test}
\end{figure}

In \Cref{fig:test} we illustrate the effects of select structured and unstructured Dropout approaches on an image of a cat. As can be seen in \Cref{fig:orignal} the original image consists of three channels (RGB) which are aggregated to form the image. Different approaches to Dropout may treat channels differently. In \Cref{fig:dropout} we illustrate the effect of unstructured Dropout on this image, the many small black squares represent deactivated/dropped weights at a pixel level and we also see different pixels have been deactivated in each channel. In \Cref{fig:dropblock} we see fewer but larger black squares, and that the locations of dropped pixels are consistent between channels, however, this is not the case in \Cref{fig:adaptive} and \Cref{fig:channel}. In this work, we say that BatchDropBlock is channel consistent i.e. channels do not deactivate blocks independently rather the deactivated blocks are consistent between channels.

In \Cref{fig:dropout}, \Cref{fig:dropblock}, \Cref{fig:channel}, for a single channel there is a uniform probability of any pixel or block (depending on the approach) to be dropped and so deactivated pixels may not contain any of the key information required to identify this image as a cat (i.e. the probability of deactivating a pixel/block belonging to the cat is the same as that of one belonging to the background). This is not the case for \Cref{fig:adaptive}, in our adaptive DropBlock approach the probability of a block being dropped is dependent on the value of the center pixel in the block. It can be seen that this approach is not channel consistent and deactivated pixels are concentrated on the cat.

\Cref{fig:test} is illustrative to give one an intuitive understanding of these techniques, as in practice these techniques are applied to feature maps which are the output activations of a preceding layer of the network. The contributions of this paper include:

\begin{itemize}
    \item The testing of preexisting unstructured and structured Dropout approaches on current state-of-the-art models including transformer-based models on natural language inference and vision tasks. We reveal that structured Dropouts are generally better than unstructured ones on both vision and language tasks.
    \item The proposal of a new approach to structured dropout named ProbDropBlock, which improved model performance on both vision and language tasks.
    ProbDropBlock is adaptive and the blocks dropped are dependent on the relative per-pixel values. It improves RoBERTa finetuning on MNLI by $0.22\%$ and ResNet50 on ImageNet by $0.28\%$.
    \item Further observation of the benefits of simple linear scheduling observed \cite{ghiasi2018dropblock} for both structured and unstructured Dropout on a range of vision and language models.
\end{itemize}





	\section{Related Work}
In this section, we briefly review related works in the areas of structured and unstructured Dropouts used as both a regularization technique to improve model performance and as an approach to pruning to reduce the model size and computational requirements. We briefly detail unstructured Dropout and the various structured Dropouts devised for other network architectures.

\subsection{Unstructured Dropout}
To help address the problem of overfitting in neural networks, \citet{nitish2014dropout} proposed Dropout as a simple way of limiting the co-adaptation of the activation of units in the network. 
By randomly deactivating units during training they sample from an exponential number of different thinned networks and at test time an ensemble of these thinned networks is approximated by a single full network with smaller weights. Dropout led to improvements in the performance of neural networks on various tasks and has become widely adopted. This form of Dropout in this work we refer to as unstructured Dropout as any combination of units in the network may be randomly dropped/deactivated. In the following subsection, we consider forms of structured Dropout which extend this idea further for other network architectures and tasks.

\subsection{Dropblock and other structured Dropouts}
\citet{ghiasi2018dropblock} proposed DropBlock as a way to perform Structured Dropout for Convolutional Neural Nets (CNNs). They suggest that unstructured Dropout is less effective for convolutional layers than fully connected layers because activation units in convolutional layers are spatially correlated so information can still flow through convolutional networks despite Dropout and so they devised DropBlock which drops units in a contiguous area of the feature map collectively. This approach was inspired by \citet{devries2017cutout}'s Cutout, a data augmentation method where parts of the input examples are zeroed out. DropBlock generalized Cutout by applying Cutout at every feature map in convolutional networks. \citet{ghiasi2018dropblock} also found that a scheduling scheme of linearly increasing DropBlock's zero-out ratio performed better than a fixed ratio.

\citet{dai2019batch} extended DropBlock to Batch DropBlock. Their network consists of two branches; a global branch and a feature-dropping branch. In their feature dropping branch they randomly zero out the same contiguous area from each feature map in a batch involved in computing loss function. They suggest zeroing out the same block in each batch allows the network to learn a more comprehensive and spatially distributed feature representation.

\citet{larsson2016fractalnet} proposed DropPath in their work on FractalNets. Just as Dropout prevents the co-adaptation of activations, DropPath prevents the co-adaptation of parallel paths in networks such as FractalNets by randomly dropping operands of the join layers. DropPath provides at least one such path while sampling a subnetwork with many other paths disabled. DropPath during training alternates between a global sampling strategy which returns only a single path and a local sampling strategy in which a join drops each input with fixed probability, but with a guarantee, at least one survives. This encourages the development of individual columns as performant stand-alone subnetworks.

\citet{cai2019effective} proposed DropConv2d as they suggest the failure of standard dropout is due to conflict between the stochasticity of unstructured dropout and the following Batch Normalization (BN) step. They propose placing dropout operations right before the convolutional operation instead of BN or replacing BN with Group Normalization (GN) to reduce this conflict. Additionally, they devised DropConv2d which draws inspiration from DropPath and DropChannel, they treat each channel connection as a path between input and output channels and perform dropout on replicates of each of these paths.

DropBlock, BatchDropBlock, DropPath and DropConv2d are forms of structured Dropout designed with specific architecture in mind. However, as seen by \citet{cai2019effective} DropConv2d an approach to structured Dropout designed for a given network can still be useful to novel network architecture. Aside from being used to improve generalization, structured Dropout has also been used as an approach to pruning and reducing computational resource requirements at inference time. 

\citet{fan2019reducing} proposed LayerDrop as a means of regularization for transformers during training and efficient pruning at inference time reducing the large amount of computation these models require. This approach is a form of structured Dropout where instead of deactivating weights independently throughout the network, weights that collectively form a single structure in the network are deactivated. Attention heads in a transformer are typically computed in parallel, as a result in the paper the structure they focused on deactivating where fully connected layers. Using this approach they found that they were able to select sub-networks of any depth from one large network that without finetuning achieved similar performance.

Other forms of run-time structured pruning include \citep{xin2020deebert, gao2018dynamic, wu2018blockdrop,wang2019pay}, \citet{xin2020deebert} proposed DeeBERT which accelerates inferencing in BERT models by allowing samples to exit earlier without passing through the entire model under certain conditions as they believe that, for BERT, features provided by the intermediate transformer layers may suffice to classify some input samples. \citet{gao2018dynamic} preserve the full network structure of CNNs and accelerates convolution by dynamically skipping unimportant input and output channels determined by a saliency criterion exploiting the fact that the importance of features computed by convolutional layers is highly input-dependent. \citet{wu2018blockdrop} propose BlockDrop where an RL agent learns which blocks in a ResNet to select dynamically for a given novel input image.
In the meantime, fine-grained pruning methods achieve good compression rates \citep{han2015deep}, but may contain ineffectual computation that is hard to utilize on commondity hardware \citep{nikolic2019characterizing}.

In this work, we are interested in dropping structured patterns in the spatial dimension and would like to understand how this style of structured Dropouts can affect current mainstream models such as Transformers and Vision Transformers. We mainly compare our proposed approach with DropBlock and a revised version of BatchDropBlock, since these two methods are dropping at the same granularity as us.
	\section{Method}
In this section we detail the approach to structured dropout we employed in this paper. Our approach is a form of adaptive DropBlock we call ProbDropBlock, it is inspired by DropBlock as it also randomly removes larger blocks from each feature map, but rather than assigning a uniform probability to each element of being the center point of the block it assigns higher probability to elements with higher activation values. 

As such the block removed depends on the model's learned representation of the feature map and we believe this encourages the model to learn a more balanced and diverse feature map. We also employ a simple linear schedule where we linearly increase the base probability $\alpha$ of dropping a block. \citet{ghiasi2018dropblock} made an observation in their paper that this approach can significantly improve performance and is more robust.

   

\begin{algorithm}

\caption{Adaptive DropBlock} \label{algo:pdb}
\begin{algorithmic}[1]
\Procedure{ProbDropBlock}{$A, B, \alpha, mode$}       
    \State \textbf{Inputs} Layer Output Activations - $A$, Block Size - $B$, Base Drop Probability - $\alpha$, $mode$

    \If{$\textit{mode} == \textit{Inference}$}
        \State \textbf{return} $A$
    \Else
        \State $\gamma_{i,j} = \dfrac{\norm{A}_{0} \times abs(A_{i,j}))}{\norm{A}_{1}}$ \Comment{Compute drop ratio $\gamma_{i,j}$ for each element in A}
        \State $q_{i,j} = \min ((\alpha \times \gamma_{i,j}), 1)$ \Comment{Compute drop probabilities $q_{i,j}$}
        \State $M : M_{i,j} \sim Bernoulli(1 - q_{i,j})$ \Comment{Randomly sample mask $M$} 
        \For{$M_{i,j}$ in $M$}
            \If{$M_{i,j} == 0$}
                \State \textbf{store}(($i$, $j$)) \Comment{Store mask indices with zero entries}
            \EndIf
        \EndFor
        \State $\beta_{lb} = floor\left(\dfrac{B-1}{2}\right)$ \Comment{Compute lower bound buffer zone for mask}
        \State $\beta_{ub} = round \left(\dfrac{B-1}{2}\right)$ \Comment{Compute upper bound buffer zone for mask}
        \For{$(i,j)$ in \textbf{store}}
            \State $M_{i-\beta_{lb}:i+\beta_{ub},j-\beta_{lb}:j+\beta_{ub}} .= 0$ \Comment{Set values in square centered at $M_{i,j}$ to 0}
        \EndFor
        \State $A = A \times M$ \Comment{Apply mask $M$ to $A$}
        \State $A = A \times \dfrac{\textbf{sum}(M)}{\norm{M}_{0}}$ \Comment{Normalize}
        \State \textbf{return} $A$
    \EndIf 
\EndProcedure

\end{algorithmic}
\end{algorithm}

In \Cref{algo:pdb} we detail our adaptive DropBlock method, the algorithm takes as input the output activations from a layer $A$, a block size $B$, base drop probability $\alpha$ and the $mode$ of the network. 

When not in inference mode the algorithm computes a drop ratio $\gamma_{i,j}$ for each entry in the feature map. This is a normalized drop ratio as illustrated in Line 6 of \Cref{algo:pdb}: this drop ratio is equal to ratio of the absolute value of the entry $abs(A_{i,j})$ and the average of absolute values of all entries in the feature map $\dfrac{\norm{A}_{0}}{\norm{A}_{1}}$. 

Each drop ratio $\gamma_{i,j}$ is multiplied with the base drop probability and constrained to range $[0,1]$ to give the drop probabilities $q_{i,j}$ for each entry in the feature map as illustrated in Line 7 of \Cref{algo:pdb}. As a result, entries with values that are higher than the mean absolute entry value for the feature map have a higher probability of being dropped. 

Using drop probabilities $q_{i,j}$ we sample a mask $M$, we modify the mask by constructing a block of size $B$ around each zero entry in the mask and setting values in the box to zero to create a larger contiguous block of zeros. We handle even block sizes by effectively shifting the block's center point by half an entry right. 

After modification this mask $M$ is applied to the feature map $A$ by element wise matrix multiplication. Finally, we re-normalize $A$ by the ratio of non-zero entries in $M$ to the total number of entries in $M$ and return it, this is the same re-normalization technique used in DropBlock \citep{ghiasi2018dropblock}. The linear scheduling is implemented by gradually increasing the base drop probability $\alpha$ over the training cycle. The details and effect of the linear scheduling are evaluated in details in \Cref{sec:eval:prob}.
	\section{Evaluation}

In the following section, we lay out the experimental setup for this paper and briefly discuss the datasets and models considered in \Cref{sec:eval:setup}. 

We then address the importance of probability scheduling for both structured and unstructured Dropout methods in \Cref{sec:eval:prob}. In \Cref{sec:eval:structure}, we demonstrate that structured Dropouts generally outperform their unstructured counterpart for both vision and language tasks.

To properly assess our approach to other structured Dropouts we test on both vision and language tasks for state-of-the-art models in \Cref{sec:eval:language} and \Cref{sec:eval:vision}. 
We compare the performance of models trained with the proposed ProbDropBlock approach to those trained with other forms of Dropout and baseline models trained without any forms of Dropout.

\subsection{Experiment Setup}
\label{sec:eval:setup}
In this work, we considered 6 datasets, 3 NLP (natural language processing) datasets and 3 CV (computer vision) datasets. 

The NLP datasets considered are all part of \textbf{GLUE} - the multitask benchmark and analysis platform for natural language understanding \citep{GLUE}, we consider MNLI, QNLI and RTE in our evaluation.

For the CV datasets, we consider CIFAR10, CIFAR100 \citep{cifar} and ImageNet \citep{deng2009imagenet} classification.
More details about these datasets can be found in \Cref{sec:appendix:datasets}.

For finetuning the RoBERTa model, we use the hyperparameter setup in \citet{liu2019roberta}. For training the ResNet family models, we train all models using the Adam optimizer \cite{kingma2014adam} and pick the best learning rate from $\{1e^{-5}, 5e^{-5}, 1e^{-4}\}$. For the pyramid vision transformer (PVT-V2) model, we use the standard setup described in \citet{wang2022pvt}. We slightly changed the augmentations in the original PVT-V2 setup, removing Mixup and Random Erasing, so that it is more closely aligned with the ResNet training setup for a better comparison.  
In the ResNet family and PVT-V2 models, we insert the Dropout mechanism after each residual block. 
For RoBERTa, we add Dropout, DropBlock or ProbDropblock to the end of each encoder layer. 

We run each data point $3$ times with different random seeds, and report both the average and standard deviations, the details of our hardware system setup can be found in \Cref{sec:appendix:hardware}.
We picked the dropping Block Size ($B$) to be $B=4$, and show an ablation of this parameter in \Cref{sec:appendix:block}.


\subsection{Probability scheduling}
\label{sec:eval:prob}
\begin{table*}[!h]
	\caption{Structured (BatchDropBlock) and non-structured Dropouts on CIFAR-10. with and without the linear scheduling on the dropping probability. BDB is BatchDropBlock. $\Delta$ shows the difference between the accuracy compared to baseline, the baseline accuracy on this task is $94.37\%$.}
	\centering
	\begin{tabular}{|c|cccc|}
	\hline 
	Method       
	& Dropout 					& Dropout-Schedule
	& BDB 				& BDB-Schedule 	 \\
	\hline
	Resnet50 Acc $\uparrow$
	& $94.50 \pm 0.04$	& $94.70 \pm 0.14$                           
	& $93.11 \pm 0.22$	& $94.77 \pm 0.29$	 \\
	$\Delta \uparrow$
	& $+0.13$						& $+0.33$                           
	& $-1.26$						& $\mathbf{+0.41}$	 \\
	\hline
	\end{tabular}
	\label{tab:schedule}
\end{table*}

\citeauthor{ghiasi2018dropblock} mentioned in their experiments that DropBlock with a linear dropping scheme that decreases the value of keep probability from $1$ to $1-\alpha$ can significantly improve the performance. We test $\alpha \in \{0.1, 0.2, 0.3, 0.5\}$, and pick the best performing $\alpha$ ($\alpha=0.2$ in this case). A detailed explanation of how we pick the keep probability is in \Cref{sec:appendix:prob}. 

In \Cref{tab:schedule}, we apply this linear dropping scheme to both the standard Dropout \citep{srivastava2014dropout} and BatchDropBlock \citep{dai2019batch}.
We consider a modified version of BatchDropBlock, where all channels of a single input are dropped consistently, but datapoints in a batch can drop independently.
One major observation from \Cref{tab:schedule} is that an appropriate probability scheduling improves the performance of both structured and unstructured Dropout methods. The scheduling shows a greater impact on structured Dropout.

Intuitively, Dropout servers as a regularization method, and applying it at the start of the training interferes with the optimization; this type of regularization method should be introduced at a later stage of training when the training accuracy starts to become larger than the validation accuracy, or in other words when overfitting starts to arise.

In general, we observed that:
\begin{itemize}
	\item Structured Dropout (\textit{eg.} DropBlock) with a linear dropping scheme of decreasing the value of keep probability can significantly improve the performance, this aligns with the observation made by \citeauthor{ghiasi2018dropblock}.
	\item \textbf{\emph{Non-structured Dropout also benefits from the linear dropping scheme}}, although it is a less significant improvement than the structured Dropout.
\end{itemize}

\subsection{Structured and non-structured Dropouts}
\label{sec:eval:structure}

In this section, we compare the performance of the standard Dropout, various structured Dropout schemes (BatchDropBlock and DropBlock) and ProbDropBlock. All methods have a linear dropping scheme of decreasing the value of the keep probability from $1$ to $1 - \alpha$. We experimented also $\alpha \in \{0.1, 0.2, 0.3, 0.5\}$, and used $\alpha=0.2$ for ResNet and $0.1$ for RoBERTa models, an ablation study of different dropping probabilities can be found in \Cref{sec:appendix:prob}.

\begin{table*}[!h]
	\caption{A comparison of the performance of Dropout and various Structured Dropout schemes. BDB is BatchDropBlock. DropBlock is not channel consistent (blocks deactivated are not consistent between channels), ProbDropBlock additionally has dropping probabilities correlated to pixel-wise saliency.
	RoBERTa is evaluated on MNLI with a baseline accuracy of $87.60\%$, and ResNet50 is evaluated on CIFAR10 with a baseline accuracy of $94.37\%$. $\Delta$ is the difference between the current accuracy and baseline.
	}
	\centering
	\begin{tabular}{|c|cccc|}
	\hline 
	Method       
	& Dropout 					
	& BDB 				
	& DropBlock 	
	& ProbDropBlock \\
	\hline
	Resnet50 Acc $\uparrow$
	& $94.70 \pm 0.14$                           
	& $94.77 \pm 0.29$	& $95.05 \pm 0.21$	& $94.73 \pm 0.19$ \\
	$\Delta \uparrow$
	& $+0.33$                           
	& $+0.41$						& $\mathbf{+0.68}$	& $\mathit{+0.35}$ \\
	\hline
	RoBERTa Acc $\uparrow$
	& $87.51 \pm 0.08$         
	& $87.39 \pm 0.29$  & $87.71 \pm 0.24$  & $87.83 \pm 0.15$ \\
	$\Delta \uparrow$
	& $-0.09$         
	& $-0.21$  							& $\mathit{+0.11}$	& $\mathbf{+0.22}$ \\
	\hline
	\end{tabular}
	\label{tab:comp}
\end{table*}

Language models such as BERT \citep{devlin2018bert} and RoBERTa \citep{liu2019roberta} are based on the multi-head attention mechanism \citep{vaswani2017attention}. 
Prior work has not studied how to apply coarse-grained Dropout techniques on transformer-based architectures.

We apply these structured Dropouts in a head-wise manner, this means for BatchDropBlock, we drop the same pattern across heads in multi-head attention. DropBlock, in contrast, then drops each head independently. 

The original RoBERTa used a Dropout with $\alpha=0.1$, and we replaced all of these Dropouts with the regularization strategies shown in \Cref{tab:comp}.
Notice, in this case, our baseline considered is a standard RoBERTa without any Dropouts. Our RoBERTa baseline on MNLI achieves $87.60\%$. The baseline accuracy for ResNet50 on CIFAR10 is $94.37\%$.

The ReNet50 model is evaluated on CIFAR10. The striding of the network is adjusted to fit into this smaller image size of CIFAR10. The details of this network architecture are summarized in \Cref{sec:appendix:networks}.
\Cref{tab:comp} confirmed with the observation made by \citet{ghiasi2018dropblock} that structured Dropouts are generally better than standard, unstructured Dropouts on vision tasks. In addition, our results in \Cref{tab:comp} also suggest that structured Dropout is better on MNLI. 

Another interesting observation is that unstructured Dropout does not provide any performance gains for language models ($-0.09$ on RoBERTa). In the meantime, we see that BDB, although works reasonably well on ResNet50, has a detrimental impact on the performance of RoBERTa. This means that applying structured Dropout methods to each Transformer head independently is important for these methods to improve the performance of language models. We will investigate this phenomenon in greater detail in \Cref{sec:eval:language}.

In general, we observed that: 
\begin{itemize}
	\item In addition to what was originally shown by \citeauthor{ghiasi2018dropblock}, 	
	we observed \textbf{\emph{structured Dropout techniques are generally better not only on vision tasks but also on language tasks.}}
	\item Structured Dropout techniques are more advantageous on language tasks compared to vision tasks.
\end{itemize}

\subsection{Language tasks}
\label{sec:eval:language}
\Cref{tab:language} demonstrates the performance of RoBERTa finetuned on three GLUE tasks (MNLI, QNLI, RTE) using different structured Dropout techniques. To our best knowledge, we are the first to investigate the effect of block-wise structured Dropout methods on Transformer-based models.

We observe that both DropBlock and ProbDropBlock consistently outperform BatchDropBlock (BDB) and that ProbDropBlock is the best performing method. Both DropBlock and ProbDropBlock can significantly help RoBERTa achieve better performance, while BDB has a negative impact on its accuracy.

The major difference with BDB is that each head drops blocks with the same pattern across heads, we observe this is having a negative effect on the performance of RoBERTa. 

The proposed strategy, ProbDropBlock, is able to achieve the best performance on all tasks and is able to outperform the original RoBERTa model by a significant margin. This is a clear indication that dropping structured patterns based on the saliency values of the attention maps is advantageous.

\begin{table}[htbp]
	\centering
	\caption{Different Structured Dropout schemes. DropBlock is channel independent (channels are not dropped independently), ProbDropBlock additionally has dropping probabilities correlated to element-wise saliency. The RoBERTa model is first pretrained on a large unlabeled text corpus and subsequently finetuned on these tasks, which is the same setup in \citeauthor{liu2019roberta}.}
	\begin{tabular}{|c|c|ccc|}
		\hline
		Method 												&	Metric		& MNLI 										& QNLI									& RTE \\ \hline
		Baseline											&	Accuracy $\uparrow$	& $87.60 \pm 0.04$  			& $92.75 \pm 0.03$ 			& $73.28 \pm 0.02$ \\
		\hline
		\multirow{2}{*}{BatchDropBlock} 				
		&	Accuracy $\uparrow$											& $87.39 \pm 0.29$  			& $92.70 \pm 0.06$  		& $70.64 \pm 0.07$ \\
		&	$\Delta$ $\uparrow$											& $-0.21$  								& $-0.05$								& $-2.64$ \\ 
		\hline
		\multirow{2}{*}{DropBlock}
		& Accuracy $\uparrow$	 										& $87.71 \pm 0.24$  			& $92.81 \pm 0.11$ 			& $72.51 \pm 0.08$ \\ 
		&	$\Delta$ $\uparrow$											& $+0.11$  								& $+0.06$								& $-0.77$ \\ 
		\hline
		\multirow{2}{*}{ProbDropBlock}
		& Accuracy $\uparrow$											& $87.83 \pm 0.15$ 				& $92.90 \pm 0.12$			& $74.25 \pm 0.03$ \\ 
		&	$\Delta$ $\uparrow$											& $\mathbf{+0.22}$  			& $\mathbf{+0.25}$			& $\mathbf{+0.97}$ \\ 
		\hline
	\end{tabular}
	\label{tab:language}
\end{table}

In general, we made the following observations from \Cref{tab:language}:
\begin{itemize}
    \item BatchDropBlock generally has a negative impact on the performance of language models.
    \item Structured Dropout applied identically per head (BatchDropBlock) does not improve model performance on language tasks, both DropBlock and ProbDropBlock outperform BDB by a significant margin. \textbf{\emph{Dropping heads independently is critical for better performance on language models.}} 
    \item The multi-head attention modules in the transformer benefit the most from ProbDropBlock.
\end{itemize}

\subsection{Vision tasks}
\label{sec:eval:vision}

\begin{table}[!ht]
	\centering
	\caption{Different Structured Dropout schemes. DropBlock is channel independent (channels are not dropped independently), ProbDropBlock additionally has dropping probabilities correlated to element-wise saliency. ResNet50 and WideResNet28 are from the ResNet family with adjusted striding to match the CIFAR image size. PVTv2-B1 is the pyramid vision transformer.}
	\label{fig:cifar}
	\begin{tabular}{|c|c|cccc|}
		\hline
		\multirow{2}{*}{Method}
		&\multirow{2}{*}{Metric}
		&\multicolumn{2}{c}{CIFAR10}
		&\multicolumn{2}{c|}{CIFAR100}
		\\
		&
		& ResNet50	
		& PVTv2-B1
		& WideResNet28
		& PVTv2-B1
		\\ \hline
		Baseline
		& Accuracy $\uparrow$
		& $94.37 \pm 0.32$
		& $95.59 \pm 1.00$
		& $74.72 \pm 0.08$
		& $82.38 \pm 0.19$
		\\
		\hline
		\multirow{2}{*}{BatchDropBlock}
		& Accuracy $\uparrow$
		& $94.77 \pm 0.29$
		& $95.99 \pm 0.15$
		& $74.97 \pm 0.26$
		& $82.22 \pm 0.34$
		\\
		& $\Delta \uparrow$
		& $+0.41$
		& $+0.40$
		& $+0.25$
		& $-0.16$
		\\
		\hline
		\multirow{2}{*}{DropBlock}
		& Accuracy $\uparrow$
		& $95.05 \pm 0.21$
		& $95.89 \pm 0.09$
		& $74.99 \pm 0.08$
		& $82.26 \pm 0.35$
		\\
		& $\Delta \uparrow$
		& $\mathbf{+0.68}$
		& $+0.30$
		& $+0.27$
		& $-0.12$
		\\
		\hline
		\multirow{2}{*}{ProbDropBlock}
		& Accuracy $\uparrow$
		& $94.73 \pm 0.19$
		& $96.15 \pm 0.01$
		& $75.13 \pm 0.27$
		& $82.44 \pm 0.16$
		\\
		& $\Delta \uparrow$
		& $+0.35$
		& $\mathbf{+0.56}$
		& $\mathbf{+0.41}$
		& $+\mathbf{0.06}$
		\\
		\hline

	\end{tabular}
\end{table}

\begin{table}[htbp]
	\centering
	\caption{ProbDropBlock and baseline performance for ImageNet classification.} 
    \label{fig:imagnet}
	\begin{tabular}{|c|c|cc|}
		\hline
		\multirow{2}{*}{Method}
		&\multirow{2}{*}{Metric}
		&\multicolumn{2}{c|}{ImageNet} \\
		&
		& ResNet50	
		& PVTv2-B1
		\\ \hline
		Baseline
		& Accuracy $\uparrow$
		& $74.22 \pm 0.06$	
		& $78.27 \pm 0.04$ \\
		\hline
		\multirow{2}{*}{ProbDropBlock}
		& Accuracy $\uparrow$
		& $74.50 \pm 0.17$	
		& $78.88 \pm 0.22$ \\
		& $\Delta \uparrow$
		& $+0.28$	
		& $+0.61$ \\
		\hline

	\end{tabular}
\end{table}

\Cref{fig:cifar} and \Cref{fig:imagnet} demonstrate the results of applying different structured Dropout methods on CIFAR10, CIFAR100 and ImageNet. Vision Transformers (ViTs) recently have demonstrated great capabilities on major vision benchmarks \citep{liu2021swin, wang2022pvt}, so we consider both the ResNet family \citep{he2016deep,zagoruyko2016wide} and Pyramid Vision Transformer \cite{wang2022pvt} in our experiment. Prior research has hardly systematically studied the effect of structured Dropouts on Vision Transformer models.

We observe that in general ProbDropBlock shows the best performance on all dataset network combinations, except for one outlier which is ResNet50 on CIFAR10. Interestingly, we observe a phenomenon that BDB (BatchDropBlock) in general improves the performance of both CNNs and ViTs according to \Cref{fig:cifar}. This is very different from the phenomenon observed in \Cref{sec:eval:language} that BDB generally decreases the performance of language models.

We also notice that only ProbDropBlock can slightly increase the accuracy of PVTv2-B1 on CIFAR100. We then realised that PVTv2-B1 has a validation accuracy that is not greatly larger than its training accuracy, meaning that this model does not overfit the dataset by a significant margin. For instance, ResNet50 on CIFAR10 overfits heavily on the CIFAR10 task and thus it benefits the most from regularization methods. The PVTv2-B1 model architecture used for both CIFAR10 and CIFAR100 is significantly smaller than the original model, the details of this model architecture difference are explained in \Cref{sec:appendix:networks}. 
In addition, CIFAR100 is a harder task than CIFAR10, we see PVTv2-B1 benefits less from regularization methods such as structured Dropouts when the model is not overfitting.

\Cref{fig:cifar} generally demonstrate that ProbDropBlock is the best regularization method on three out of the four model-dataset combinations.
When we train PVTv2-B1 on CIFAR100, both BDB and DropBlock fail to improve the model performance since the model does not overfit greatly to the dataset; but, ProbDropBlock still makes a positive impact on model performance. 
We further tested the effect of ProbDropBlock on ImageNet and \Cref{fig:imagnet} summarizes our results. We demonstrate that ProbDropBlock is an effective regularization method, it provides us $+0.28\%$ and $+0.61\%$ accuracy gains for ResNet50 and PVTv2-B1 respectively.

Structured Dropout techniques in general can help vision models based on our observations on \Cref{fig:cifar} and \Cref{fig:imagnet}. We make the following observations:
\begin{itemize}
    \item BatchDropBlock has a positive impact on computer vision models, this is different from what we have observed in \Cref{sec:eval:language}.
    \item Models that are not overfitting benefit less from structured Dropouts.
    \item Structured Dropout methods generally help vision models to learn better, \textbf{\emph{the proposed ProbDropBlock is effective on both CNNs and ViTs.}}
\end{itemize}



	\section{Conclusion}
In this paper, we revisit the ideas of structured Dropout for current state-of-the-art models, we devise our form of adaptive structured Dropout - ProbDropBlock and compare preexisting structured and unstructured Dropout approaches to ours on vision and language tasks. We demonstrated the utility of a simple linear dropping schedule for both structured and unstructured Dropouts supporting a similar observation made by \citet{ghiasi2018dropblock}. 

Our approach, ProbDropBlock was able to achieve improvements in performance for all networks and task combinations and outperformed other forms of Dropout considered for the majority of combinations we evaluated. In particular, ProbDropBlock improved RoBERTa finetuning on MNLI by $0.22\%$, and training of ResNet50 on ImageNet by $0.28\%$. 

This work demonstrates the utility of structured Dropout approaches not just on residual networks and CNNs, but on language and vision transformers. However, there is a limit to the gains achievable through Dropout alone, as demonstrated by the results of the PVTv2-B1 model on the CIFAR-100 dataset, when there is minimal overfitting in the model regularization only provides minimal gain. 
	\newpage
	\section{Ethics Statement}
Language and vision models typically require a significant amount of data for training and testing. Our work only uses datasets that are widely used within the ML community, however, there are some ethical concerns about the collection methodologies and entries in some of these datasets \cite{DBLP:journals/corr/abs-2109-13228}. We make use of these datasets because they are established and recognized benchmarks while we acknowledge possible ethical issues with these datasets. In this work, we focus on improving generalization by combating overfitting through adaptive structured Dropout. Generally, improved model generalization is a positive outcome but that assumes these trained models are not applied to nefarious purposes, however, we can not ensure this.
	\section{Reproducibility Statement}
We discuss how we setup our experiments in \Cref{sec:eval:setup}. We explained the learning rate and optimizer setup in that section. At the beginning of \Cref{sec:eval:prob}, \Cref{sec:eval:language} and \Cref{sec:eval:vision}, we explained the choice of our probability scheduling and the picked $\alpha$ value. In \Cref{sec:appendix:prob} and \Cref{sec:appendix:block}, we explained how we picked the hyperparameters $\alpha$ and $B$ based on these ablation studies. Each of our experiments is repeated for $3$ times, and both the arithmetic mean and standard deviation are reported. 

\Cref{sec:appendix:datasets} details the setup of each dataset and \Cref{sec:appendix:networks} details the models we have used and included any modifications we have made to these models. Our hardware system used to perform experiments is reported in \Cref{sec:appendix:hardware}.
	\appendix
\section{Datasets}
\label{sec:appendix:datasets}
\subsection{Language Tasks}
MNLI (Multi-Genre Natural Language Inference) corpus \citep{N18-1101,GLUE} is a collection of sentence pairs with textual entailment annotations gathered via crowd sourcing. The sentences are paired as premise and hypothesis and the task is to predict if the premise entails the hypothesis (\textit{entailment}), contradicts the hypothesis (\textit{contradiction}) or neither (\textit{neutral}). The corpus is modeled on the SNLI (Stanford Natural Lanfuage Inference) corpus \citep{bowman2015large}, but differs in that covers a range of genres of spoken and written text, and supports a distinctive cross-genre generalization evaluation \citep{N18-1101,GLUE}. The premises are gathered from ten different sources, including fiction, government reports and transcribed speeches. It consists of 393k train samples and 20k test samples.

QNLI (Question-answering Natural Language Inference) corpus is a dataset automatically derived from SQuAD (Stanford Question Answering Dataset) \citep{rajpurkar2016squad, GLUE}. SQuAD is a question-answering dataset which consists of question-paragraph pairs, where a sentence in the paragraph contains the answer to the corresponding question. QNLI is constructed by converting the task into sentence pair classification by forming a pair between each question and each sentence in the corresponding context. It consists of 105k training samples and 5.4k testing samples.

\subsubsubsection{RTE}
RTE The Recognizing Textual Entailment (RTE) datasets is a combination of several datasets which came from a series of annual textual entailment challenges \cite{GLUE,dagan2005pascal,haim2006second,giampiccolo2007third,bentivogli2009fifth}. It consists of 2.5k training samples and 3k testing samples.




\subsection{Vision Tasks}
\subsubsubsection{CIFAR-10 \& CIFAR-100}
The CIFAR-10 dataset contains 60000 32x32 colour images divided into 10 classes, each with 6000 images. There are 50,000 training and 10,000 test images.

The CIFAR-100 dataset is just like the CIFAR-10, except it has 100 classes containing 600 images each. There are 500 training images and 100 testing images per class. The 100 classes in the CIFAR-100 are grouped into 20 superclasses. Each image comes with a "fine" label (the class to which it belongs) and a "coarse" label (the superclass to which it belongs).

The ILSVRC 2012 image classification dataset contains 1.2 million images for training and 50,000 for validation from 1000 classes. The input image sizes are $224 \times 224$ center crop to images at test time. The results are reported on the validation set.



\section{Networks}
\label{sec:appendix:networks}
\subsection{RoBERTa}
The RoBERTa model is kept the same as its original form proposed by \citet{liu2019roberta}.
In our experiment, we consider the RoBERTa-base model only.
The base model contains $12$ layers, with a hidden size of $768$, an FFN inner hidden size of $3072$ and $12$ attention heads.
The original model uses Dropout and we replace all of the original Dropouts to structured Dropout methods.

\subsubsection{ResNet}
We consider both the original ResNet \citep{he2016deep} and its wider alternative (WideResNet) \citep{zagoruyko2016wide}.
These networks normally have one convolutional layer (named stem) and four other residual blocks. For the CIFAR10 and CIFAR100 classification, we change the striding of the first convolution to $1$ and deleted the first max pooling. These adaptions help the network to operate with the $32 \times 32$ image size on CIFAR datasets

\subsubsection{PVT-V2}
\Cref{tab:pvt_setup} demonstrates the detailed setup of the PVT-V2 structure used for CIFAR and ImageNet tasks. The rest of the setup parameters are the same as \citet{wang2022pvt}, and the setup is the same as the PVTV2-B1 model.
\begin{table}[htbp]
	\centering
	\caption{PVT-V2 setup for vision datasets, $e$ is the embedding dimension and $s$ is the striding used for the overlapping patch embedding.} 
    \label{tab:pvt_setup}
	\begin{tabular}{|c|l l|}
		\hline
		Layer name
		& CIFAR10/CIFAR1000
		& ImageNet 
		\\ \hline
		Stage 1
		& $e=16, s=4$
		& $e=64, s=4$ \\
		\hline
		Stage 2
		& $e=32, s=2$
		& $e=128, s=2$ \\
		\hline
		Stage 3
		& $e=64, s=1$
		& $e=256, s=2$ \\
		\hline
		Stage 2
		& $e=128, s=2$
		& $e=512, s=2$ \\
		\hline
	\end{tabular}
\end{table}

\section{Hardware System}
\label{sec:appendix:hardware}
We used a variety of hardware systems, our initial testing and CIFAR10 results are generated on a hardware system with 4 x NVIDIA GeForce RTX 2080 Ti GPUs. The ImageNet training and RoBERTa training are performed on 4 x Nvidia A100 SXM4 80GB GPUs. The total amount of GPU training cost for all the expriments in this paper is around 20 GPU-days.

\section{Picking the dropping probability}

\begin{table}[htbp]
	\centering
	\caption{Different Dropping probabilities for ProbDropBlock.}
	\label{fig:block_size}
	\begin{tabular}{|c|ccc|}
		\hline
		Probability
		&ResNet50 on CIFAR10
		&PVT-V2 on CIFAR100
		&RoBERTa on MNLI 
		\\ \hline
		0.0
		& $94.37 \pm 0.32$
		& $82.38 \pm 0.19$
		& $87.60 \pm 0.04$
		\\ 
		0.1
		& $94.70 \pm 0.14$
		& $\mathbf{82.44 \pm 0.16}$
		& $\mathbf{87.83 \pm 0.15}$
		\\ 
		0.2
		& $\mathbf{94.73 \pm 0.19}$	
		& $82.21 \pm 0.13$
		& $87.32 \pm 0.11$
		\\
		0.3
		& $94.20 \pm 0.30$	
		& $82.10 \pm 0.16$
		& $69.45 \pm 0.16$
		\\
		\hline
	\end{tabular}
\end{table}

\label{sec:appendix:prob}

\section{Picking the block size}
\label{sec:appendix:block}
The block size for DropBlock is a hyperparameter that needs to be tuned. 
We test $B \in \{2, 4, 6, 8, 10\}$, and pick the best performing $B$ 
($B=4$ in this case).
\begin{figure}[!h]
		\centering
		\includegraphics[scale=0.5]{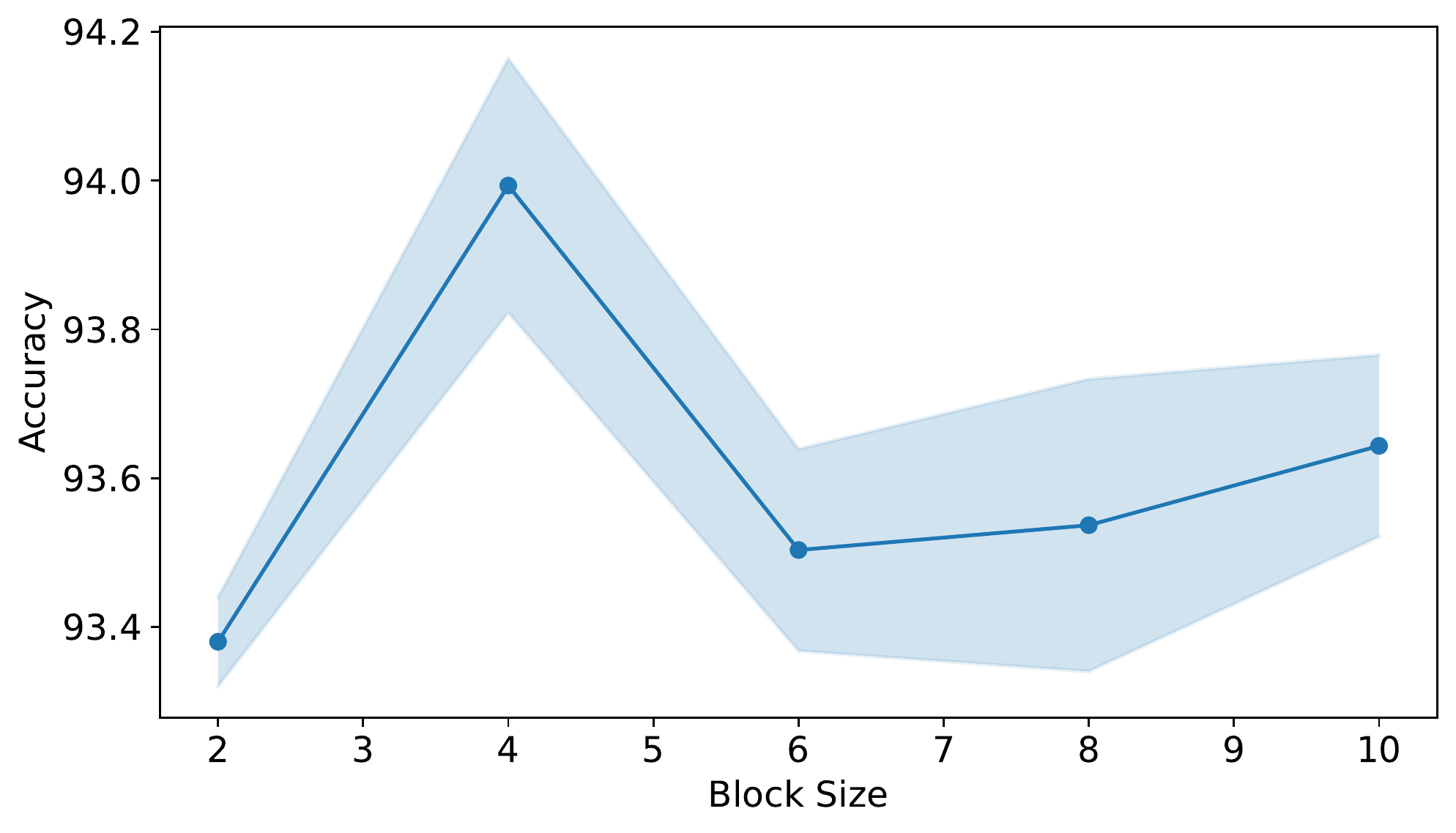}
		\caption{The effect of block size on the performance of DropBlock.}
		\label{fig:block_size}	
\end{figure}

\bibliographystyle{unsrtnat}  
\bibliography{paper_arxiv}

\begin{thebibliography}{40}
\providecommand{\natexlab}[1]{#1}
\providecommand{\url}[1]{\texttt{#1}}
\expandafter\ifx\csname urlstyle\endcsname\relax
  \providecommand{\doi}[1]{doi: #1}\else
  \providecommand{\doi}{doi: \begingroup \urlstyle{rm}\Url}\fi

\bibitem[Heaton(2020)]{heaton2020applications}
Jeff Heaton.
\newblock Applications of deep neural networks.
\newblock \emph{arXiv preprint arXiv:2009.05673}, 2020.

\bibitem[Jumper et~al.(2021)Jumper, Evans, Pritzel, Green, Figurnov,
  Ronneberger, Tunyasuvunakool, Bates, {\v{Z}}{\'\i}dek, Potapenko,
  et~al.]{jumper2021highly}
John Jumper, Richard Evans, Alexander Pritzel, Tim Green, Michael Figurnov,
  Olaf Ronneberger, Kathryn Tunyasuvunakool, Russ Bates, Augustin
  {\v{Z}}{\'\i}dek, Anna Potapenko, et~al.
\newblock Highly accurate protein structure prediction with alphafold.
\newblock \emph{Nature}, 596\penalty0 (7873):\penalty0 583--589, 2021.

\bibitem[Schrittwieser et~al.(2020)Schrittwieser, Antonoglou, Hubert, Simonyan,
  Sifre, Schmitt, Guez, Lockhart, Hassabis, Graepel,
  et~al.]{schrittwieser2020mastering}
Julian Schrittwieser, Ioannis Antonoglou, Thomas Hubert, Karen Simonyan,
  Laurent Sifre, Simon Schmitt, Arthur Guez, Edward Lockhart, Demis Hassabis,
  Thore Graepel, et~al.
\newblock Mastering atari, go, chess and shogi by planning with a learned
  model.
\newblock \emph{Nature}, 588\penalty0 (7839):\penalty0 604--609, 2020.

\bibitem[Caruana et~al.(2000)Caruana, Lawrence, and
  Giles]{caruana2000overfitting}
Rich Caruana, Steve Lawrence, and C~Giles.
\newblock Overfitting in neural nets: Backpropagation, conjugate gradient, and
  early stopping.
\newblock \emph{Advances in neural information processing systems}, 13, 2000.

\bibitem[DeVries and Taylor(2017)]{devries2017improved}
Terrance DeVries and Graham~W Taylor.
\newblock Improved regularization of convolutional neural networks with cutout.
\newblock \emph{arXiv preprint arXiv:1708.04552}, 2017.

\bibitem[Loshchilov and Hutter(2017)]{loshchilov2017decoupled}
Ilya Loshchilov and Frank Hutter.
\newblock Decoupled weight decay regularization.
\newblock \emph{arXiv preprint arXiv:1711.05101}, 2017.

\bibitem[Srivastava et~al.(2014{\natexlab{a}})Srivastava, Hinton, Krizhevsky,
  Sutskever, and Salakhutdinov]{srivastava2014dropout}
Nitish Srivastava, Geoffrey Hinton, Alex Krizhevsky, Ilya Sutskever, and Ruslan
  Salakhutdinov.
\newblock Dropout: a simple way to prevent neural networks from overfitting.
\newblock \emph{The journal of machine learning research}, 15\penalty0
  (1):\penalty0 1929--1958, 2014{\natexlab{a}}.

\bibitem[He et~al.(2016)He, Zhang, Ren, and Sun]{he2016deep}
Kaiming He, Xiangyu Zhang, Shaoqing Ren, and Jian Sun.
\newblock Deep residual learning for image recognition.
\newblock In \emph{Proceedings of the IEEE conference on computer vision and
  pattern recognition}, pages 770--778, 2016.

\bibitem[Huang et~al.(2017)Huang, Liu, Van Der~Maaten, and
  Weinberger]{huang2017densely}
Gao Huang, Zhuang Liu, Laurens Van Der~Maaten, and Kilian~Q Weinberger.
\newblock Densely connected convolutional networks.
\newblock In \emph{Proceedings of the IEEE conference on computer vision and
  pattern recognition}, pages 4700--4708, 2017.

\bibitem[Ghiasi et~al.(2018)Ghiasi, Lin, and Le]{ghiasi2018dropblock}
Golnaz Ghiasi, Tsung-Yi Lin, and Quoc~V Le.
\newblock Dropblock: A regularization method for convolutional networks.
\newblock \emph{Advances in neural information processing systems}, 31, 2018.

\bibitem[Dai et~al.(2019)Dai, Chen, Gu, Zhu, and Tan]{dai2019batch}
Zuozhuo Dai, Mingqiang Chen, Xiaodong Gu, Siyu Zhu, and Ping Tan.
\newblock Batch dropblock network for person re-identification and beyond.
\newblock In \emph{Proceedings of the IEEE/CVF international conference on
  computer vision}, pages 3691--3701, 2019.

\bibitem[Cai et~al.(2019)Cai, Shu, Chen, Ooi, Wang, and
  Zhang]{cai2019effective}
Shaofeng Cai, Yao Shu, Gang Chen, Beng~Chin Ooi, Wei Wang, and Meihui Zhang.
\newblock Effective and efficient dropout for deep convolutional neural
  networks.
\newblock \emph{arXiv preprint arXiv:1904.03392}, 2019.

\bibitem[Xin et~al.(2020)Xin, Tang, Lee, Yu, and Lin]{xin2020deebert}
Ji~Xin, Raphael Tang, Jaejun Lee, Yaoliang Yu, and Jimmy Lin.
\newblock Deebert: Dynamic early exiting for accelerating bert inference.
\newblock \emph{arXiv preprint arXiv:2004.12993}, 2020.

\bibitem[Fan et~al.(2019)Fan, Grave, and Joulin]{fan2019reducing}
Angela Fan, Edouard Grave, and Armand Joulin.
\newblock Reducing transformer depth on demand with structured dropout.
\newblock \emph{arXiv preprint arXiv:1909.11556}, 2019.

\bibitem[Srivastava et~al.(2014{\natexlab{b}})Srivastava, Hinton, Krizhevsky,
  Sutskever, and Salakhutdinov]{nitish2014dropout}
Nitish Srivastava, Geoffrey Hinton, Alex Krizhevsky, Ilya Sutskever, and Ruslan
  Salakhutdinov.
\newblock Dropout: A simple way to prevent neural networks from overfitting.
\newblock \emph{Journal of Machine Learning Research}, 15\penalty0
  (56):\penalty0 1929--1958, 2014{\natexlab{b}}.
\newblock URL \url{http://jmlr.org/papers/v15/srivastava14a.html}.

\bibitem[Devries and Taylor(2017)]{devries2017cutout}
Terrance Devries and Graham~W. Taylor.
\newblock Improved regularization of convolutional neural networks with cutout.
\newblock \emph{CoRR}, abs/1708.04552, 2017.
\newblock URL \url{http://arxiv.org/abs/1708.04552}.

\bibitem[Larsson et~al.(2016)Larsson, Maire, and
  Shakhnarovich]{larsson2016fractalnet}
Gustav Larsson, Michael Maire, and Gregory Shakhnarovich.
\newblock Fractalnet: Ultra-deep neural networks without residuals.
\newblock \emph{arXiv preprint arXiv:1605.07648}, 2016.

\bibitem[Gao et~al.(2018)Gao, Zhao, Dudziak, Mullins, and Xu]{gao2018dynamic}
Xitong Gao, Yiren Zhao, {\L}ukasz Dudziak, Robert Mullins, and Cheng-zhong Xu.
\newblock Dynamic channel pruning: Feature boosting and suppression.
\newblock \emph{arXiv preprint arXiv:1810.05331}, 2018.

\bibitem[Wu et~al.(2018)Wu, Nagarajan, Kumar, Rennie, Davis, Grauman, and
  Feris]{wu2018blockdrop}
Zuxuan Wu, Tushar Nagarajan, Abhishek Kumar, Steven Rennie, Larry~S Davis,
  Kristen Grauman, and Rogerio Feris.
\newblock Blockdrop: Dynamic inference paths in residual networks.
\newblock In \emph{Proceedings of the IEEE conference on computer vision and
  pattern recognition}, pages 8817--8826, 2018.

\bibitem[Wang et~al.(2019)Wang, Gao, Zhao, Li, Dou, and Xu]{wang2019pay}
Kafeng Wang, Xitong Gao, Yiren Zhao, Xingjian Li, Dejing Dou, and Cheng-Zhong
  Xu.
\newblock Pay attention to features, transfer learn faster cnns.
\newblock In \emph{International conference on learning representations}, 2019.

\bibitem[Han et~al.(2015)Han, Mao, and Dally]{han2015deep}
Song Han, Huizi Mao, and William~J Dally.
\newblock Deep compression: Compressing deep neural networks with pruning,
  trained quantization and huffman coding.
\newblock \emph{arXiv preprint arXiv:1510.00149}, 2015.

\bibitem[Nikoli{\'c} et~al.(2019)Nikoli{\'c}, Mahmoud, Moshovos, Zhao, and
  Mullins]{nikolic2019characterizing}
Milo{\v{s}} Nikoli{\'c}, Mostafa Mahmoud, Andreas Moshovos, Yiren Zhao, and
  Robert Mullins.
\newblock Characterizing sources of ineffectual computations in deep learning
  networks.
\newblock In \emph{2019 IEEE International Symposium on Performance Analysis of
  Systems and Software (ISPASS)}, pages 165--176. IEEE, 2019.

\bibitem[Wang et~al.(2018)Wang, Singh, Michael, Hill, Levy, and Bowman]{GLUE}
Alex Wang, Amanpreet Singh, Julian Michael, Felix Hill, Omer Levy, and
  Samuel~R. Bowman.
\newblock {GLUE:} {A} multi-task benchmark and analysis platform for natural
  language understanding.
\newblock \emph{CoRR}, abs/1804.07461, 2018.
\newblock URL \url{http://arxiv.org/abs/1804.07461}.

\bibitem[cif()]{cifar}
{Cifar-10} and {cifar-100} datasets.
\newblock \url{https://www.cs.toronto.edu/~kriz/cifar.html}.
\newblock Accessed: 2022-09-22.

\bibitem[Deng et~al.(2009)Deng, Dong, Socher, Li, Li, and
  Fei-Fei]{deng2009imagenet}
Jia Deng, Wei Dong, Richard Socher, Li-Jia Li, Kai Li, and Li~Fei-Fei.
\newblock Imagenet: A large-scale hierarchical image database.
\newblock In \emph{2009 IEEE conference on computer vision and pattern
  recognition}, pages 248--255. Ieee, 2009.

\bibitem[Liu et~al.(2019)Liu, Ott, Goyal, Du, Joshi, Chen, Levy, Lewis,
  Zettlemoyer, and Stoyanov]{liu2019roberta}
Yinhan Liu, Myle Ott, Naman Goyal, Jingfei Du, Mandar Joshi, Danqi Chen, Omer
  Levy, Mike Lewis, Luke Zettlemoyer, and Veselin Stoyanov.
\newblock Roberta: A robustly optimized bert pretraining approach.
\newblock \emph{arXiv preprint arXiv:1907.11692}, 2019.

\bibitem[Kingma and Ba(2014)]{kingma2014adam}
Diederik~P Kingma and Jimmy Ba.
\newblock Adam: A method for stochastic optimization.
\newblock \emph{arXiv preprint arXiv:1412.6980}, 2014.

\bibitem[Wang et~al.(2022)Wang, Xie, Li, Fan, Song, Liang, Lu, Luo, and
  Shao]{wang2022pvt}
Wenhai Wang, Enze Xie, Xiang Li, Deng-Ping Fan, Kaitao Song, Ding Liang, Tong
  Lu, Ping Luo, and Ling Shao.
\newblock Pvt v2: Improved baselines with pyramid vision transformer.
\newblock \emph{Computational Visual Media}, 8\penalty0 (3):\penalty0 415--424,
  2022.

\bibitem[Devlin et~al.(2018)Devlin, Chang, Lee, and Toutanova]{devlin2018bert}
Jacob Devlin, Ming-Wei Chang, Kenton Lee, and Kristina Toutanova.
\newblock Bert: Pre-training of deep bidirectional transformers for language
  understanding.
\newblock \emph{arXiv preprint arXiv:1810.04805}, 2018.

\bibitem[Vaswani et~al.(2017)Vaswani, Shazeer, Parmar, Uszkoreit, Jones, Gomez,
  Kaiser, and Polosukhin]{vaswani2017attention}
Ashish Vaswani, Noam Shazeer, Niki Parmar, Jakob Uszkoreit, Llion Jones,
  Aidan~N Gomez, {\L}ukasz Kaiser, and Illia Polosukhin.
\newblock Attention is all you need.
\newblock \emph{Advances in neural information processing systems}, 30, 2017.

\bibitem[Liu et~al.(2021)Liu, Lin, Cao, Hu, Wei, Zhang, Lin, and
  Guo]{liu2021swin}
Ze~Liu, Yutong Lin, Yue Cao, Han Hu, Yixuan Wei, Zheng Zhang, Stephen Lin, and
  Baining Guo.
\newblock Swin transformer: Hierarchical vision transformer using shifted
  windows.
\newblock In \emph{Proceedings of the IEEE/CVF International Conference on
  Computer Vision}, pages 10012--10022, 2021.

\bibitem[Zagoruyko and Komodakis(2016)]{zagoruyko2016wide}
Sergey Zagoruyko and Nikos Komodakis.
\newblock Wide residual networks.
\newblock \emph{arXiv preprint arXiv:1605.07146}, 2016.

\bibitem[Asano et~al.(2021)Asano, Rupprecht, Zisserman, and
  Vedaldi]{DBLP:journals/corr/abs-2109-13228}
Yuki~Markus Asano, Christian Rupprecht, Andrew Zisserman, and Andrea Vedaldi.
\newblock {PASS:} an imagenet replacement for self-supervised pretraining
  without humans.
\newblock \emph{CoRR}, abs/2109.13228, 2021.
\newblock URL \url{https://arxiv.org/abs/2109.13228}.

\bibitem[Williams et~al.(2018)Williams, Nangia, and Bowman]{N18-1101}
Adina Williams, Nikita Nangia, and Samuel Bowman.
\newblock A broad-coverage challenge corpus for sentence understanding through
  inference.
\newblock In \emph{Proceedings of the 2018 Conference of the North American
  Chapter of the Association for Computational Linguistics: Human Language
  Technologies, Volume 1 (Long Papers)}, pages 1112--1122. Association for
  Computational Linguistics, 2018.
\newblock URL \url{http://aclweb.org/anthology/N18-1101}.

\bibitem[Bowman et~al.(2015)Bowman, Angeli, Potts, and
  Manning]{bowman2015large}
Samuel~R. Bowman, Gabor Angeli, Christopher Potts, and Christopher~D. Manning.
\newblock A large annotated corpus for learning natural language inference.
\newblock In \emph{Proceedings of the 2015 Conference on Empirical Methods in
  Natural Language Processing (EMNLP)}. Association for Computational
  Linguistics, 2015.

\bibitem[Rajpurkar et~al.(2016)Rajpurkar, Zhang, Lopyrev, and
  Liang]{rajpurkar2016squad}
Pranav Rajpurkar, Jian Zhang, Konstantin Lopyrev, and Percy Liang.
\newblock Squad: 100,000+ questions for machine comprehension of text.
\newblock \emph{arXiv preprint arXiv:1606.05250}, 2016.

\bibitem[Dagan et~al.(2005)Dagan, Glickman, and Magnini]{dagan2005pascal}
Ido Dagan, Oren Glickman, and Bernardo Magnini.
\newblock The pascal recognising textual entailment challenge.
\newblock In \emph{Machine learning challenges workshop}, pages 177--190.
  Springer, 2005.

\bibitem[Haim et~al.(2006)Haim, Dagan, Dolan, Ferro, Giampiccolo, Magnini, and
  Szpektor]{haim2006second}
R~Bar Haim, Ido Dagan, Bill Dolan, Lisa Ferro, Danilo Giampiccolo, Bernardo
  Magnini, and Idan Szpektor.
\newblock The second pascal recognising textual entailment challenge.
\newblock In \emph{Proceedings of the Second PASCAL Challenges Workshop on
  Recognising Textual Entailment}, volume~7, 2006.

\bibitem[Giampiccolo et~al.(2007)Giampiccolo, Magnini, Dagan, and
  Dolan]{giampiccolo2007third}
Danilo Giampiccolo, Bernardo Magnini, Ido Dagan, and William~B Dolan.
\newblock The third pascal recognizing textual entailment challenge.
\newblock In \emph{Proceedings of the ACL-PASCAL workshop on textual entailment
  and paraphrasing}, pages 1--9, 2007.

\bibitem[Bentivogli et~al.(2009)Bentivogli, Clark, Dagan, and
  Giampiccolo]{bentivogli2009fifth}
Luisa Bentivogli, Peter Clark, Ido Dagan, and Danilo Giampiccolo.
\newblock The fifth pascal recognizing textual entailment challenge.
\newblock In \emph{TAC}, 2009.

\end{thebibliography}

\end{document}